\documentclass[12pt,a4paper,twoside]{article}
\makeatletter\if@twocolumn\PassOptionsToPackage{switch}{lineno}\else\fi\makeatother

\usepackage{amsfonts,amssymb,amsbsy,latexsym,amsmath,tabulary,graphicx,times,xcolor}
\usepackage[T1]{fontenc}
\usepackage{url,multirow,morefloats,floatflt,cancel,tfrupee,multicol}
\makeatletter

\AtBeginDocument{\@ifpackageloaded{textcomp}{}{\usepackage{textcomp}}}
\makeatother
\usepackage{colortbl}
\usepackage{xcolor}
\usepackage{pifont}
\usepackage[nointegrals]{wasysym}
\usepackage[superscript,biblabel]{cite}
\makeatletter

\@ifundefined{etal}{\def\etal{\textit{et~al}}}{}

\AtBeginDocument{
\expandafter\ifx\csname eqalign\endcsname\relax
\def\eqalign#1{\null\vcenter{\def\\{\cr}\openup\jot\m@th
  \ialign{\strut$\displaystyle{##}$\hfil&$\displaystyle{{}##}$\hfil
      \crcr#1\crcr}}\,}
\fi
}

\@ifundefined{tsGraphicsScaleX}{\gdef\tsGraphicsScaleX{1}}{}
\@ifundefined{tsGraphicsScaleY}{\gdef\tsGraphicsScaleY{.9}}{}
\def\checkGraphicsWidth{\ifdim\Gin@nat@width>\linewidth
	\tsGraphicsScaleX\linewidth\else\Gin@nat@width\fi}

\def\checkGraphicsHeight{\ifdim\Gin@nat@height>.9\textheight
	\tsGraphicsScaleY\textheight\else\Gin@nat@height\fi}

\def\fixFloatSize#1{}%\@ifundefined{processdelayedfloats}{\setbox0=\hbox{\includegraphics{#1}}\ifnum\wd0<\columnwidth\relax\renewenvironment{figure*}{\begin{figure}}{\end{figure}}\fi}{}}
\let\ts@includegraphics\includegraphics

\def\inlinegraphic[#1]#2{{\edef\@tempa{#1}\edef\baseline@shift{\ifx\@tempa\@empty0\else#1\fi}\edef\tempZ{\the\numexpr(\numexpr(\baseline@shift*\f@size/100))}\protect\raisebox{\tempZ pt}{\ts@includegraphics{#2}}}}

\AtBeginDocument{\def\includegraphics{\@ifnextchar[{\ts@includegraphics}{\ts@includegraphics[width=\checkGraphicsWidth,height=\checkGraphicsHeight,keepaspectratio]}}}

\DeclareMathAlphabet{\mathpzc}{OT1}{pzc}{m}{it}

\def\URL#1#2{\@ifundefined{href}{#2}{\href{#1}{#2}}}

\edef\fntEncoding{\f@encoding}

\makeatother

\newif\ifmultipleabstract\multipleabstractfalse%
\usepackage[hmargin=1.8cm,vmargin=2.2cm]{geometry}%,headsep=0.3in,footskip=0.49in
\makeatletter
\def\author#1{\gdef\@author{\hskip-\dimexpr(\tabcolsep)\hskip1pt\parbox{\dimexpr\textwidth-1pt}{\centering #1}}}

\let\@articletype\@empty \def\articletype#1{\gdef\@articletype{{\fontsize{14}{16}\selectfont #1}}}

\usepackage{fancyhdr}

\fancypagestyle{headings}{\fancyhf{}\fancyhead[RE]{\MakeTextUppercase{\textbf{\RunningAuthor}}}\fancyhead[LE]{\textbf{\thepage}}\fancyhead[LO]{\MakeTextUppercase{\textbf{\RunningHead}}}\fancyhead[RO]{\textbf{\thepage}}}
\fancypagestyle{plain}{\fancyhf{}\fancyfoot[C]{\textbf{\thepage}}}

\AtBeginDocument{\usepackage{lastpage}}
\def\title#1{%
  \gdef\@title{%
    \ifx\@articletype\@empty\else\@articletype~\\\fi%
     #1}%
}

\usepackage{abstract}
\def\abstractname{\textbf{Abstract}}
\renewenvironment{onecolabstract}
{\vspace*{-.4pc}\trivlist\item[]\leftskip1pt\noindent\selectfont\hfill\abstractname\hfill\mbox{\null}\par\ignorespaces}{\endtrivlist}

\def\NormalBaseline{\def\baselinestretch{1.1}}

\usepackage{textcase}
\usepackage[explicit]{titlesec}
\titleformat{\section}[block]{\NormalBaseline\boldmath\bfseries}
{\thesection.}
{6pt}
{#1}
[]
\titleformat{\subsection}[hang]{\NormalBaseline\filright\itshape}
{\thesubsection.}
{6pt}
{#1}
[]
\titleformat{\subsubsection}[runin]{\NormalBaseline\filright\itshape}
{\hspace{16pt}\thesubsubsection}
{6pt}
{#1}
[]
\titleformat{\paragraph}[runin]{\NormalBaseline}
{\theparagraph}
{6pt}
{#1}
[]
\titleformat{\subparagraph}[runin]{\NormalBaseline}
{\thesubparagraph}
{6pt}
{#1}
[]

\titlespacing{\section}{0pt}{1.5\baselineskip}{.2\baselineskip}  
\titlespacing{\subsection}{0pt}{1.5\baselineskip}{.2\baselineskip}  
\titlespacing{\subsubsection}{0pt}{1.5\baselineskip}{.2\baselineskip}  
\titlespacing{\paragraph}{0pt}{.5\baselineskip}{10pt}  
\titlespacing{\subparagraph}{0pt}{.5\baselineskip}{10pt}

\usepackage{caption}
\DeclareCaptionLabelFormat{figlabel}{Figure #2}
\date{}
\makeatother

\usepackage{endfloat}
\usepackage{float}
\usepackage{setspace}

\begin{document}

\title{A cut-and-fold self-sustained compliant oscillator for autonomous actuation of origami-inspired robots}
\def\RunningHead{A cut-and-fold self-sustained compliant oscillator}
\def\RunningAuthor{Yan \etal}
\author{Wenzhong Yan\textsuperscript{1} and Ankur Mehta\textsuperscript{2}
\thanks{Wenzhong Yan is with Mechanical and Aerospace Engineering Department, University of California, Los Angeles, CA 90095; Ankur Mehta is with Electrical and Computer Engineering Department, University of California, Los Angeles, CA 90095. {\tt\small wzyan24@g.ucla.edu, mehtank@ucla.edu}}%
}

\maketitle

%%%%%%%%%%%%%%%%%%%%%%%%%%%%%%%%%%%%%%%%%%%%%%%%%%%%%%%%%%%%%%%%%%%%%%%%%%

{\begin{onecolabstract}
Origami-inspired robots are of particular interest given their potential for rapid and accessible design and fabrication of elegant designs and complex functionalities through cutting and folding of flexible 2D sheets or even strings, i.e.~printable manufacturing. Yet, origami robots still require bulky, rigid components or electronics for actuation and control to accomplish tasks with reliability, programmability, ability to output substantial force, and durability, restricting their full potential. 
Here, we present a printable self-sustained compliant oscillator that generates periodic actuation using only constant electrical power, without discrete components or electronic control hardware. 
This oscillator is robust (9 out of 10 prototypes worked successfully on the first try), configurable (with tunable periods from 3 s to 12 s), powerful (can overcome hydrodynamic resistance to consistently propel a swimmer at $\sim$1.6 body lengths/min), and long-lasting ($\sim$10\textsuperscript{3} cycles); it enables driving macroscale devices with prescribed autonomous behaviors, e.g. locomotion and sequencing. This oscillator is also fully functional underwater and in high magnetic fields.
Our analytical model characterizes essential parameters of the oscillation period, enabling programmable design of the oscillator. 
The printable oscillator can be integrated into origami-inspired systems seamlessly and monolithically, allowing rapid design and prototyping; the resulting integrated devices are lightweight, low-cost, compliant, electronic-free, and nonmagnetic, enabling practical applications in extreme areas. 
We demonstrate the functionalities of the oscillator with: (i) autonomous gliding of a printable swimmer, (ii) LED flashing, and (iii) fluid stirring.
% and (iv) directional crawling of an origami walker. 
This work paves the way for realizing fully printable autonomous robots with a high integration of actuation and control. 

\def\keywordstitle{Keywords}
\end{onecolabstract}}
 
%%%%%%%%%%%%%%%%%%%%%%%%%%%%%%%%%%%%%%%%%%%%%%%%%%%%%%%%%%%%%%%%%%%%%%%%%%

\begin{multicols}{2}

\section*{Introduction}
Folding of patterned flexible 2D sheet materials (or even linear stings), as a top-down, parallel transformation approach, provides an opportunity to simplify and accelerate the design and fabrication of objects to achieve elegant designs and complex functionalities \cite{rus_design_2018}. In nature, folding plays an important role in the creation of a wide spectrum of complex biological structures, such as proteins and insect wings. Inspired by nature, roboticists explore folding as a design and fabrication strategy to build origami-inspired printable robots \cite{onal_origami-inspired_2013,onal_towards_2011}. Here, we refer to ‘printable robots’ as robots created through origami-inspired cutting and folding \cite{rus_design_2018}. This printable manufacturing strategy enables constructing a wide range of robotic morphologies and functionalities, such as crawling\cite{pagano_crawling_2017}, grasping\cite{faber_bioinspired_2018}, swimming\cite{miyashita_untethered_2015}, shape morphology\cite{lee_origami_2017,liu_foldable_2020}, self-folding\cite{shin_self-assembling_2014}, locomotion\cite{mulgaonkar_flying_2016}, or combinations of these tasks\cite{chen_biologically_2017,cui_nanomagnetic_2019,ren_multi-functional_2019}. 
Such origami-inspired printable devices have several potential advantages, including rapid design and fabrication\cite{Zhakypov_millirobots_2019}, low cost and high accessibility\cite{mehta2014}, high strength-to-weight ratio\cite{kim_origami-inspired_2018}, built-in compliance for safe interaction with humans\cite{Lin_origami-skeletons_2020}, compact storage and transport\cite{hawkes_programmable_2010}, reconfigurability\cite{overvelde_rational_2017} and self-folding\cite{felton_method_2014}, and high scalablility to be still practical at micro/nano-scale\cite{xu_millimeter-scale_2019,rothemund_folding_2006}.
Despite these functional advantages, origami-inspired robots may need to rely on bulky components and electronics for actuation and control to accomplish tasks with reliability, programmability, ability to output substantial force, and durability. For example, although having origami printable bodies, many robots still use electromagnetic motors and microcontrollers for actuation and control \cite{onal_towards_2011,mehta2014,felton_method_2014}. 
Dependency on such devices restricts the full potential of origami robots, especially in applications where such components or electronics do not apply \cite{rus_design_2018}.

In soft robotics, recent advances in materials and mechanisms have enabled autonomous operations without rigid components or electronics, only using a constant power supply. Fluidic oscillators, including microfluidic logic\cite{mosadegh_integrated_2010}, soft oscillators\cite{rothemund_soft_2018}, and pneumatic ring oscillators\cite{preston_soft_2019}, are able to generate oscillatory pressure outputs without hard components or electronic control. These oscillations could be correspondingly harnessed to implement meaningful physical tasks, e.g periodic actuation\cite{wehner_integrated_2016}, earthworm-like walking\cite{rothemund_soft_2018}, and rolling locomotion\cite{preston_soft_2019} for fully soft robots only requiring a constant pressure source. 
Another class of soft robots rely on self-sustained oscillation for actuation and control to realize autonomous functionalities, e.g. propulsion and shape morphing \cite{chen_scaling_2015}. These oscillations are induced from special configurations with constant stimulus fields, including humidity gradients\cite{treml_origami_2018}, temperature discrepancies\cite{wang_-built_2018,kotikian_untethered_2019} or visible light patterns\cite{zhao_soft_2019}.
Entirely soft dielectric elastomer robots are capable of accomplishing tasks autonomously thanks to dielectric elastomer logic gates, which function similar to their electronic analogy while can exert mechanical outputs. The dielectric elastomer logic gates are based on the interaction of dielectric elastomer switch/dielectric elastomer actuator (DES/DEA) \cite{henke_soft_2017,obrien_artificial_2012}, which is high-energy-density and capable of exerting large strain actuation. However, these technologies are built around complex material fabrication processes, limiting the accessibility promised by origami-inspired manufacturing; the realization of printable autonomous actuation and control of cut-and-fold robots remains a challenge.

Here, we present a printable self-sustained compliant oscillator that generates periodic actuation without electronics or control hardware. Based on our previous proof-of-concept ``mechanical logic'' \cite{yan_iser} that demonstrated fluctuating electrical signals from a constant power supply through mechanical interactions, this work substantially improves the design and extends the concept to realize autonomous actuation for origami-inspired robots. This oscillator is robust, configurable, powerful, and long-lasting, and enables actuating macroscale devices to autonomously accomplish meaningful prescribed tasks with reliability, programmability, ability to output substantial force, and durability. The printable oscillator is built of non-rigid raw materials using origami-inspired cut-and-fold manufacturing methods; this approach enables integrated design and rapid prototyping, enhances accessibility rise through low cost materials and processes, as well as increases the potential for interaction with humans with built-in compliance.

\begin{figure*}[t!]
\centering
\includegraphics[trim=1.8in 8.0cm 1.98in 2.7cm, clip=true,width=11cm]{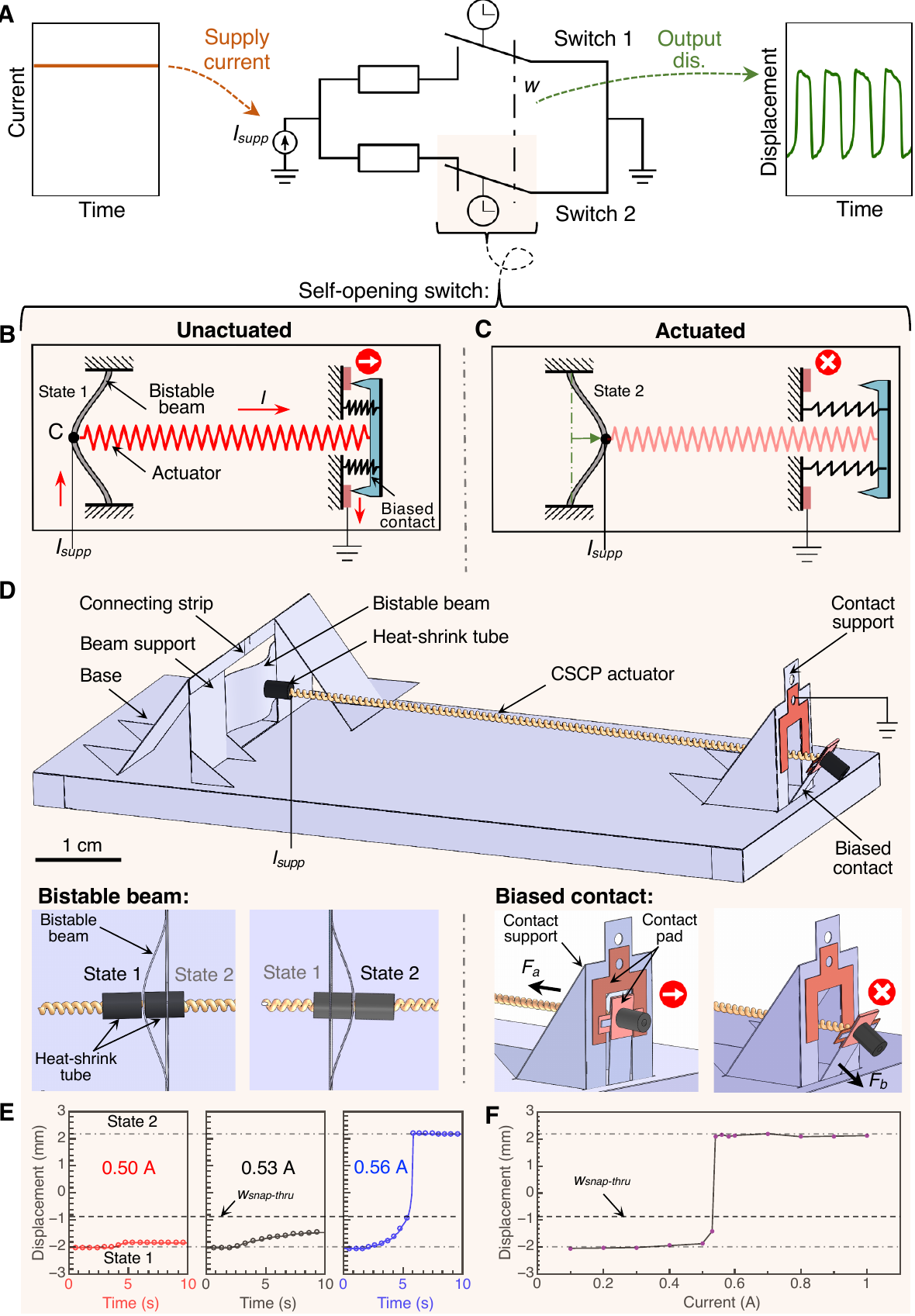}
\caption{A printable, self-sustained compliant oscillator composed of two interconnected self-opening swit\-ches. \textit{(A)} A printable oscillator is composed of two switches in parallel with their poles linked. This special configuration leads to a systematic instability of any pole position: a constant electric current supply, $I_{supp}$, drives an oscillatory displacement output $w$ of the poles. The top views of the switch show its circuit and schematic, with the unactuated state (biased contact closed, switch ``on") in \textit{(B)} and the actuated state (biased contact opened, switch ``off") in \textit{(C)}. \textit{(D)} The detailed structure of a self-opening switch. The two ends of the CSCP actuator are connected with the bistable beam and biased contact, respectively. Both bistable beam and biased contact have two different states (bottom). When the bistable beam is at state 1, the pulling force $F_{a}$ of the actuator keeps the contact closed; otherwise the contact stays opened by the bias force, $F_{b}$. \textit{(E)} Time-resolved displacement of the centerpoint C of the bistable beam under supply currents of 0.50 A, 0.53 A, and 0.56 A, respectively. \textit{(F)} Stable displacement of centerpoint C after actuation as a function of the supply current.}
\label{fig:1}
\end{figure*}

The fundamental unit of our oscillator is a printable, self-opening switch that disconnects its circuit after a geometrically programmed timed delay. The self-opening function is realized by integrating a bistable buckled beam and a conductive super coiled polymer (CSCP) actuator \cite{yip_control_2017}---a conductive thermal actuator, functionally similar to a coiled shape memory alloy (SMA) wire.  The actuator gradually contracts due to Joule heating from a constant power supply and drives the bistable beam snap through to the other state leading to the opening of the circuit. 
The oscillator is composed of two mechanically coupled self-opening switches; the oscillation results from the system-level coupling of the opening of one switch to the closing of the other ensuring no stable state can exist in this design (Fig. \ref{fig:1}A). The oscillation period of the printable oscillator depends on the self-opening switch's timed delay, which can be characterized as a function of its material properties, geometric design parameters, and supply current. To enable systematic design of the printable oscillator, we present an analytical model for the oscillation period as a function of those three categories of parameters. 
We demonstrate the capabilities of our printable oscillator by (i) autonomous gliding of a printable swimmer on water, (ii) animating an LED display, and (iii) mixing of fluid.
Specifically, these resulting devices and robots are fully printable, lightweight, low-cost, compliant, electronic-free, and nonmagnetic, allowing for extensive practical applications, especially in extreme areas.

\section*{Design}
\subsection*{The printable, self-opening switch}
The key mechanism generating a self-sustained oscillator is a printable self-opening switch. This self-opening switch provides a method to realize a timed delay effect in printable structures, which can lead to more sophisticated control and logic functions when appropriately integrated. This switch derives its operation from a precompressed bistable buckled beam with two stable equilibrium states coupled to a CSCP actuator (see Supplementary Fig. S2 for detailed fabrication process). It is worth noting that the performance of CSCP actuators in isolation has been well characterized and modeled in \cite{yip_control_2017}. We observed the same response in our experiments so that we can directly apply the knowledge into our study (see Supplementary Fig. S12 for detailed characteristics). When supplied with an electrical current as presented in Fig. \ref{fig:1}B and C, this switch functions as a normally-closed timer \cite{horowitz_art_2015}. Initially, the switch is in the unactuated state with the biased contact closed (Fig. \ref{fig:1}B).  The actuator gradually increases its temperature by Joule heating, resulting in an applied force on the bistable beam. When the displacement of the beam exceeds a given displacement threshold $w_{snap-thru}$, the beam experiences snap-through, triggering a rapid transition between bistable states and leading to the opening of the electrical connection of the biased contact (Fig. \ref{fig:1}C). Therefore, this is a self-opening switch characterized by a timed delay. Potentially, this switch could be desired as an actuator in applications where controllable actuation duration or instantaneous large geometrical changes are required\cite{wani_light-driven_2017,baumgartner_lesson_nodate}. Before starting the next operation, the switch needs to be reset with the actuator cooled down to environment temperature and the bistable beam toggled back to the initial state.
The operating frequency of this switch is typically limited by the (passive) cooling time $T_{cool}$ that is needed to reset the CSCP actuator since the (driven) heating of the actuator could be much faster. 

The self-opening switch is composed of a bistable buckled beam, a CSCP actuator, and a biased contact connected in series (Fig. \ref{fig:1}D, with detailed fabrication process in Supplementary Note 1). One end of the actuator is connected to the biased contact, whose other terminal is electrically grounded; the other end is attached on the bistable beam and connected to a current supply, $I_{supp}$. Due to the axial displacement constraints, the originally straight beam decreases in length by being compressed; it then buckles into a cosine-wave shape after reaching a compression threshold, resulting in a bistability. The beam does not require power to remain in either unactuated or actuated states, although switching between states does require energy. When the beam, at state 1, is driven rightwards by the actuator to a critical displacement, $w_{snap-thru}$, it will snap to its state 2 (Fig. \ref{fig:1}C); it can be reset to state 1 (Fig. \ref{fig:1}B) with a critical displacement leftwards, $w_{snap-back}$.

The delay of the switch is determined by the time needed for the actuator to heat up sufficiently to initiate the bistable beam's snap-through (Supplementary Fig. S1). To demonstrate the self-opening mechanism, we built a prototype device. As shown in Fig. \ref{fig:1}E, the bistable beam cannot reach the threshold displacement at low supply currents---the resulting equilibrium temperature and thus drive force in the CSCP actuator is too low. Once the drive current surpassed a threshold, the actuator was then able to drive the beam pass its threshold causing snap-through into state 2 (Fig. \ref{fig:1}F, Supplementary Movie S1). 
Increasing the supply current further beyond this threshold reduces the time needed for the actuator to heat up to its critical temperature and thus the time delay of the switch, increasing the operating speed of the device. In conclusion, the self-opening switch features a configurable timed delay effect and can serve as one-shot actuators with instantaneous large geometrical changes.

\subsection*{The printable, self-sustained oscillator}
The printable oscillator consists of two aforementioned self-opening switches connected mechanically and electrically (Fig. \ref{fig:1}A). The two switches are connected electrically in parallel (i.e., with the same current supply connections).
To generate oscillation, we mechanically linked the two switches' poles---coupling opposing unactuated and actuated states respectively---forming a double pole, single throw switch. To physically implement this mechanical linkage, the two switches can be integrated by sharing the same bistable beam leading to a simpler configuration (Fig. \ref{fig:2}A, see Supplementary Note 1 for fabrication details). The motion of the timed opening of one switch will automatically reset the other (back to the unactuated state, still needing to cool down the actuator). Thus, no stable state exists for this system composed of two connected switches; a system-level instability is built up as the switches close and open sequentially and asynchronously, each shifted by 180$^{\circ}$ in phase (Fig. \ref{fig:2}, B and C). This system-level instability results in periodic  oscillation of the displacement of the centerpoint C of the bistable beam with a minimal unstable transition period (Fig. \ref{fig:2}D) \cite{preston_soft_2019}.

\begin{figure*}[t!]
% \begin{figure*}[ht]
  \centering
  \includegraphics[trim=1.9in 9.3cm 1.8in 2.7cm, clip=true,width=11cm]{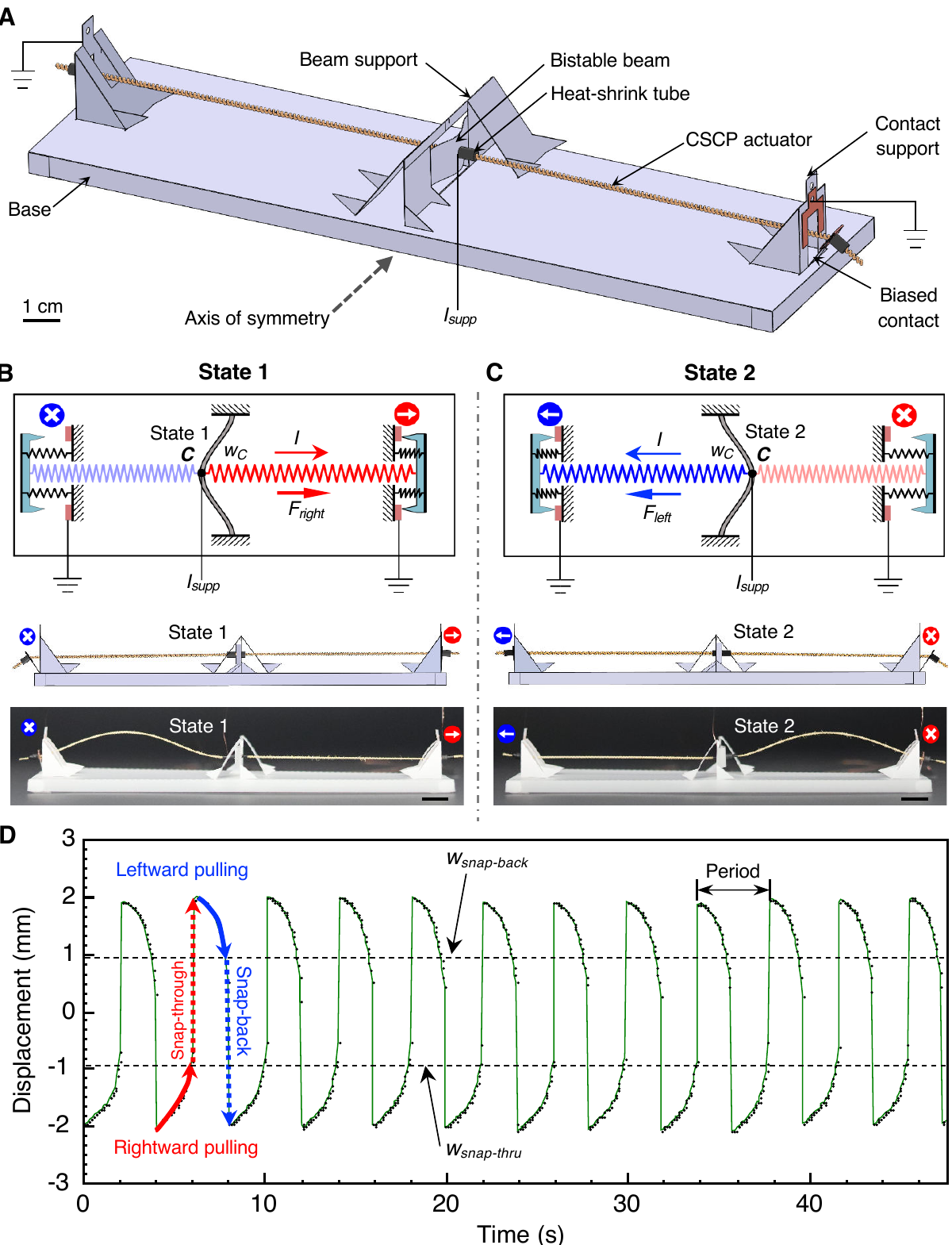}
  \caption{Operation of a printable self-sustained oscillator. \textit{(A)} The oscillator is created by combining two switches in a head-to-head configuration with a shared bistable beam; the actuated state of one switch is coupled to the unactuated state of the other \textit{(B)} and vice versa \textit{(C)}. Correspondingly, these two state are labeled as state 1 and state 2, respectively (top, top view of the 2D schematic; middle, side view of the 3D rendering; bottom, photo of the side view). \textit{(D)} The two-switch oscillator generates an oscillating output motion when a constant current power supply ($I_{supp}$ = 0.60 A) is applied.  $w_{snap-thru}$ is the displacement required to transition the bistable beam from its state 1 to its state 2; and $w_{snap-back}$ is the displacement needed to transition from its state 2 to its state 1. The red and blue arrows overlaid onto the plot indicate the rightward pull and leftward pulling processes, respectively. The corresponding red and blue dashed arrows represent the snap-through and snap-back motions. Scale bar, 1 cm (see Supplementary Movie S2 for complete oscillation).}
  \label{fig:2}
\end{figure*}

To demonstrate, a printable oscillator is fabricated with two integrated identically designed self-opening switches (Fig. \ref{fig:2}A). The mechanical structure is made from commercially available flexible DuraLar\textsuperscript{TM} Polyester Film and conductive yarn, with $w_{snap-thru}$, $\approx$ -0.89 mm and $w_{snap-back}$,  $\approx$ 0.87 mm (see Supplementary Note 1 and 7 for detailed fabrication and characterization processes, respectively), and supplied with a constant electric current power ($I_{supp}$ = 0.60 A). In order to reduce the reset time $T_{cool}$, a forced air source is supplied. After about 3 initial transient cycles, the oscillator started to generate a stable periodic oscillation of the output displacement of the midpoint C of the bistable beam (see Supplementary Movie S2) with oscillation period, $T_{osc}$ = 3.93 s, calculated by averaging its peak-to-peak periods in the time-displacement curve (see Fig. \ref{fig:2}D). Each oscillation consists of four phases: rightward pulling, snap-through, leftward pulling, and snap-back (Fig. \ref{fig:2}D). The snap-through and snap-back phases are significantly faster than the pulling phases: in this case the snap-through time was $T_{snap-thru}\approx$ 0.12 s (3.1\% of the period) and snap-back time was $T_{snap-back}\approx$ 0.10 s (2.5\% of the period). 

We further investigated the envelope of the printable oscillator design. We varied the input current and found out the shortest oscillation period was about 3.15 s when the current reached 0.63 A for this specific design. Beyond this value, oscillation could not be sustained as the actuators could not cool down between periods. After several oscillations, the bistable beam will stop at some position between two stable equilibrium states due to the simultaneous pulling of the two actuators. A broader operating range of the oscillation period could be obtained by simultaneously adjusting the speeds of actuation and cooling.
In addition, the oscillator was able to continuously cycle for about 2,000 complete periods without obvious change in frequency or degradation in performance; the oscillator continued but with a deviation in oscillation frequency due to the thermal instability of the actuators over an additional 8,000 periods.
We also explored the possibility of the oscillator working in different environments. Firstly, we submerged the oscillator completely under water; the oscillator was able to oscillate with a period of around 1.21 s from a 1.58 A supply (Supplementary Movie S3) due to the reduced cooling time $T_{cool}$ resulting from the higher thermal conductivity of water. This oscillator could be used as an actuator for underwater robots leading to a new application domain for origami-inspired designs. Secondly, we placed the oscillator into a strong magnetic field of 0.18 T (about 4000 times Earth's field). The oscillator was unaffected by either constant or dynamic magnetic fields (Supplementary Movie S4); this demonstrates the potential of this oscillator for applications within otherwise challenging high magnetic environments, such as Magnetic Resonance Imaging (MRI) systems (with fields of 0.2 -- 7 T). 

\section*{Analysis model for the printable oscillator}
The oscillator is a deeply coupled thermal-mechanical-electrical system. %, which makes the modeling extremely challenging. 
To describe the fundamental physics of the oscillator, we performed theoretical modeling to establish a governing formula. Due to the symmetry of the oscillator, we need only consider its behavior within one single self-opening motion of a switch. Thus, the oscillation period, $T_{osc}$, of an oscillator is two times of the sum of the pulling time, $T_{pull}$, and snap-through duration, $T_{snap}$ (Fig. \ref{fig:2}D):
% $T_{osc} = 2 (t_{pull}+t_{snap})$.
\begin{equation}
    T_{osc} = 2 (T_{pull}+T_{snap})
\label{eq:t_period_full}
\end{equation}

\noindent where $T_{pull}$ can be specifically defined as the duration that the actuator takes to drive the bistable beam to reach its snap-through point. 

Since the actuation of the CSCP actuator features a significantly larger characteristic time constant than the dynamics of the bistable beam \cite{yan_rapid_2019}, we propose a quasi-static assumption to simplify our model: in response to the force generated by the actuator, the bistable beam is able to achieve equilibrium instantly. Therefore, the dynamics of the actuator dictates the behavior of the entire system. In addition, on the scale of our consideration, the snap-through duration, $T_{snap}$, is typically rather small \cite{gomez_critical_2017}. For example, $T_{snap}$ is about 0.1 s while the pulling time is about 2.0 s (see Supplementary Movie S2, Fig. \ref{fig:2}C and Fig. \ref{fig:3}B). Therefore, we can neglect $T_{snap}$ and approximate the oscillation period of an oscillator as two times of the pulling time, i.e. $T_{osc} \approx 2T_{pull}$. Once we know the snap-through characteristics of the bistable beam, we can obtain the snap-through temperature of the actuator from its thermomechanical model \cite{yip_control_2017}, and thus calculate the time period (i.e. $T_{pull}$) that the actuator takes to reach the snap-through temperature from its first-order thermoelectric model. Therefore, the oscillation period, $T_{osc}$, then can be written as (with derivation in Supplementary Note 8):
{\small
\begin{equation}
% \resizebox{1\hsize}{!}{
\begin{split}
    T_{osc}=&-2\frac{C_{th}}{\lambda}ln\{1-\\
    &\frac{\lambda [F_{snap-thru}-k(w_{snap-thru}-w_{rise})]}{c_{T} I_{supp}^2 R}\}
\label{eq:t_period}
\end{split}
\end{equation}

% \small
\begin{equation}
% \resizebox{0.95\hsize}{!}{
    I_{supp} > \sqrt{\frac{\lambda}{c_{T}R} [F_{snap-thru}-k(w_{snap-thru} - w_{rise})]}
\label{eq:i_lower_bound}
\end{equation}
}
\noindent $I_{supp}$ is the supplied electrical current through the actuator with a lower bound for oscillation defined by Eq. \ref{eq:i_lower_bound}. 

$C_{th}$ and $\lambda$ are the thermal mass and absolute thermal conductivity of the actuator in its specific environment. $R$, $k$, and $c_{T}$ refer to the actuator's resistance, stiffness and thermal coefficient. 
It is worth noting that some parameters vary with actuator geometry: $\lambda$, $C_{th}$, and $R$ are directly proportional while $k$ is inversely proportional to the length of the CSCP actuator. $c_{T}$ is a constant of the material. 

On the other hand, $w_{rise}$, $w_{snap-thru}$, and $F_{snap-thru}$, are the critical snap-through characteristics of the bistable beam, respectively initial rise, critical displacement, and critical force (see Supplementary Fig. S13). All these three parameters are determined by the bistable beam's material, geometry, boundary conditions, and loading \cite{yan_analytical_2019}. According to Eq. \ref{eq:t_period}, $T_{osc}$ can be arbitrarily designed, under the assumptions of negligible snap-through duration.  However, its lower bound is determined by the cooling time, $T_{cool}$, that the opposite actuator needs to take to reset to the environment temperature before the next cycle, constraining $T_{pull} = T_{osc}/2 \geq T_{cool}$.

\begin{figure}[t]
  \centering
  \includegraphics[trim=2.5in 14.1cm 2.5in 2.5cm, clip=true,width=0.46\textwidth]{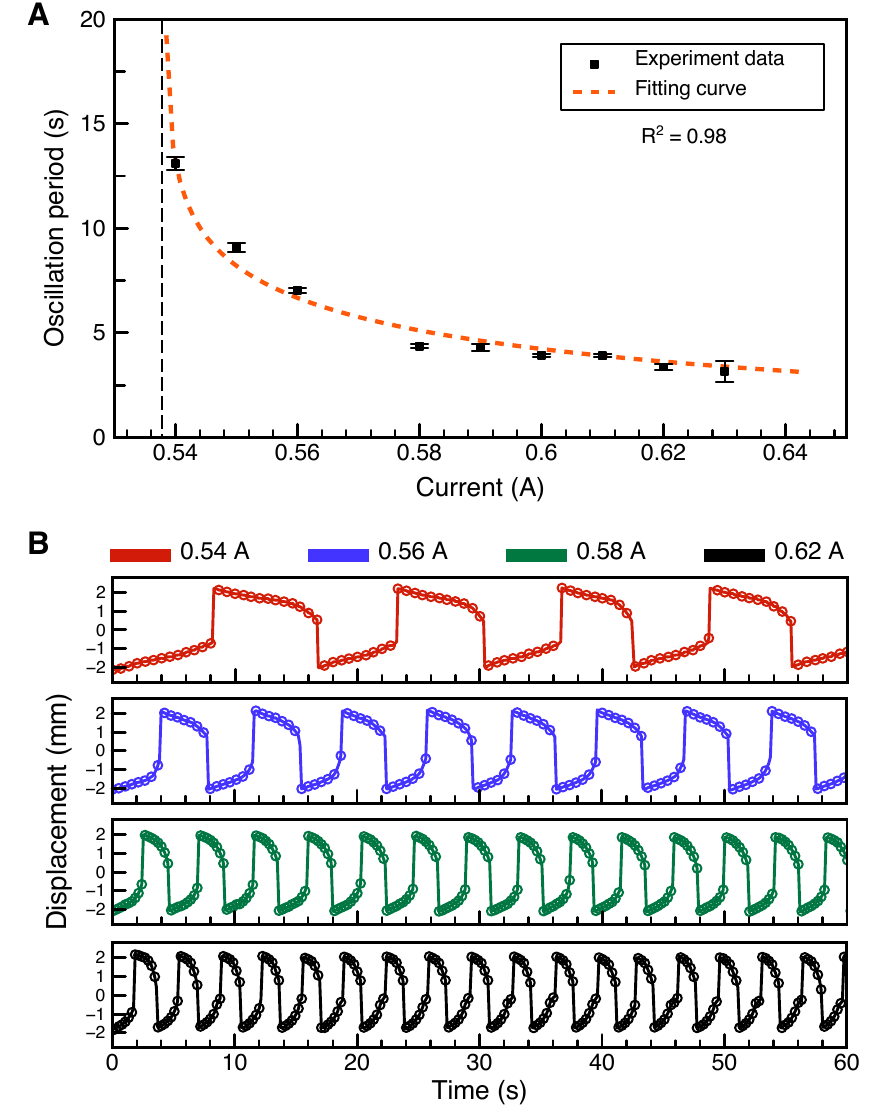}
  \caption{Oscillation period varies with the supply current. \textit{(A)} Oscillation period, $T_{osc}$, as a function of the current amplitude. The fit curve (red dashed curve, $R^2$ = 0.98) matches well with the experimental data, predicting the lower bound of the applicable current supply as around 0.538 A. Below this lower bound, the oscillator is no long capable of oscillating. The curve also indicates that the oscillation period can be arbitrarily low with sufficient current (provided equally short cooling time). Error bars represent standard deviations. \textit{(B)} Time-resolved oscillation displacement curves with various supply currents: $I_{supp}$ = 0.54 A (red); $I_{supp}$ = 0.56 A (blue), $I_{supp}$ = 0.58 A (green), and $I_{supp}$ = 0.62 A (black).}
  \label{fig:3}
\end{figure}

In our experiment, we used a 69.0 mm actuator with $R$ = 3.8 $\Omega$, $k$ = -0.28 N/mm, and $c_{T}$ = 1.6 $\times$ 10\textsuperscript{-2} N/$^{\circ}$C (see Supplementary Note 6). The bistable beam features snap-through characteristics measured to be $w_{rise}$ = -2.12 mm, $w_{snap-thru}$ = -0.89 mm, and $F_{snap-thru}$ = 0.42 N (see Supplementary Note 7). The influence of supply current on oscillation period is shown in Fig. \ref{fig:3}A. When the current increased, $T_{osc}$ dropped monotonically; the corresponding oscillation curves of the center point of the bistable beam are presented in Fig. \ref{fig:3}B. During the experiment, the oscillator was unable to oscillate outside the range of [0.54 A, 0.63 A] of the supply current. When the current is smaller than 0.54 A, the actuator could not generate enough force to activate the snap-through motion due to the insufficient equilibrium temperature. Meanwhile, oscillation is not sustainable when the $T_{osc}/2$ is smaller than $T_{cool}$ due to the switch reset requirements. By fitting the experimental results of the oscillation period versus current, we obtained the the thermal mass, $C_{th}$, and absolute thermal conductivity, $\lambda$, of the actuator to be 2.99$\times$10\textsuperscript{-2} Ws/$^{\circ}$C and 2.31$\times$10\textsuperscript{-2} W/$^{\circ}$C, respectively. The fitting curve (the red dash line in Fig. \ref{fig:3}A) indicates the lower bound of the current is around 0.538 A (from Eq.\ref{eq:i_lower_bound}), which is very close to our experimental observation, i.e. 0.54 A. In addition, the fitting curve suggests that the period asymptotically approaches 0 with increasing supply power, however, real limits on the period include cooling time, snap-through duration, and inertial dynamics outside our assumed bounds.

Though dramatically simplified, our model showed a good agreement with the experimental observation, suggesting that it may be used as an analytical tool to predict the system behaviors of the printable oscillator (and other similar devices composed of bistable mechanisms and  actuators). In addition, this analytical model can be potentially used to create a design tool for rapid prototyping of the oscillator due to its simplicity and explicitness\cite{yan_rapid_2019}. It is worth noting that the oscillation period would increase when external loading is applied on the bistable beam, with this loading effect resulting in the delay of the onset of the snap-through. However, the characterization of this effect has not been included in this paper, which may require additional analysis.

\section*{Applications of the self-sustained oscillator}
By embedding actuation and control directly into printable mechanisms, we can then build fully printable devices and robots monolithically fulfilling greater potentials of printable manufacturing. To demonstrate the abilities of our printable oscillator, we proposed four disparate applications built on our design: autonomously propelling a printable swimmer on water utilizing a rotary paddle driven by the oscillator (Fig. \ref{fig:7}), animating an LED display using the periodic motion of the oscillator to control electrical circuits (Fig. \ref{fig:5}), and mixing fluid through the oscillatory dragging of an agitator actuated by the oscillator (see Supplementary Note 9).

\subsection*{Autonomous gliding of a fully printable swimmer on water}

\begin{figure*}[t!]
  \centering
  \includegraphics[trim=0.6in 10.6cm 0.6in 2.7cm, clip=true,width=17cm]{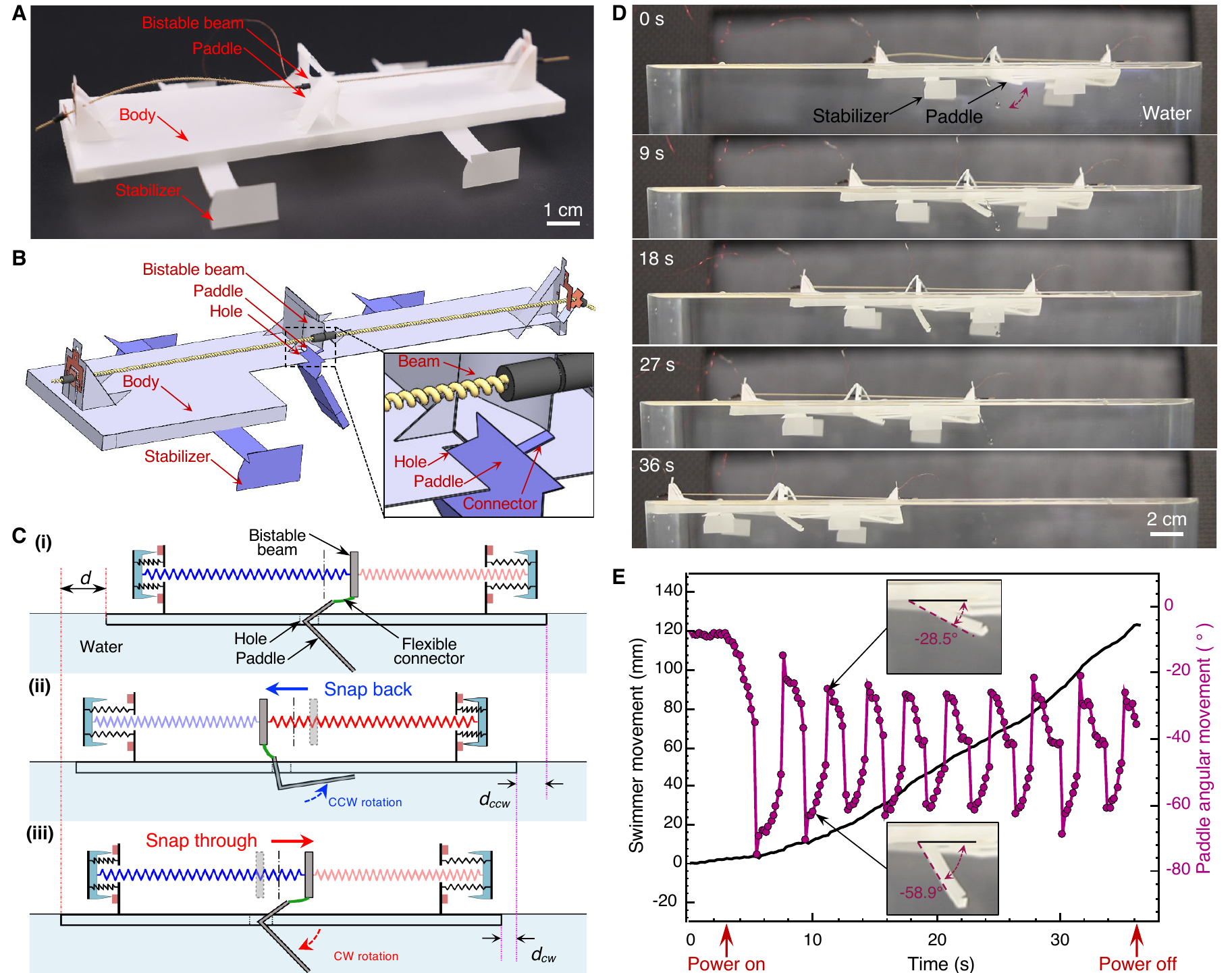}
  \caption{Autonomous gliding of a fully printable, compliant swimmer on water. \textit{(A)} The swimmer consists of an oscillator, four stabilizers, and one L-shaped paddle. The stabilizers and paddle are folded structures and can be monolithically integrated into the oscillator design. \textit{(B)} The stabilizers and paddle (in purple) are added onto the edges of the oscillator and the bottom edge of the bistable beam, respectively. The paddle is connected with the bistable beam through a flexible origami connector (see zoom-in view). \textit{(C)} Simplified locomotion mechanism of the swimmer. (i) When the bistable beam snaps back (leftward in the figure), it drives the paddle counter-clockwise, propelling the swimmer leftward by a distance, $d_{ccw}$. (ii) When the bistable beam snaps through (rightward in the figure), the paddle thrusts the swimmer move further by another distance, $d_{cw}$. Thus, the swimmer can glide leftward by $d (=d_{ccw}$ + $d_{cw})$ totally in one cycle. \textit{(D)} Power was supplied after about 3-seconds; the resulting oscillation of the bistable beam drives the rotational motion of the paddle to propel the swimmer leftward over the water surface (Supplementary Movie S5). \textit{(E)} Time-resolved plots of the (angular) displacement of the swimmer and paddle. The average speed is about 3.66 mm/s.}
  \label{fig:7}
\end{figure*}

We demonstrated the ability of the oscillator to achieve translational locomotion of origami-inspired robots by using only a constant current input. We designed a fully printable swimming robot (Fig. \ref{fig:7}A) that can glide on water surface by harnessing the periodic rotational propulsion of a paddle \cite{chen_harnessing_2018}. This rotational propulsion is generated through a mechanism that guides the paddle travel through an asymmetric stroke driven by the symmetric oscillation of the bistable beam. The swimmer is built upon the printable oscillator with additional integrated origami-inspired features (see Fig. \ref{fig:7}B and Supplementary Fig. S8A). We added four stabilizers around the oscillator and one paddle onto the bottom edge of the bistable beam (see Supplementary Note 2 for fabrication details). These stabilizers have three functions: (i) to balance the swimmer to avoid rolling; (ii) to control the gliding direction via vertical fins; (iii) to reduce friction against the sidewalls (if necessary) via horizontal bumpers. The L-shaped paddle is connected with the bistable beam through a flexible connector and constrained by a narrow hole on the body. The flexible connector joint enables the angular movement of the paddle while being compatible with the origami-inspired manufacturing method. Additional features are added onto the edges of the paddle to increase the mechanical efficiency of propulsion in water and thus, the generated thrust (Fig. S8B).

It is worth noting that all of the newly added components (including four stabilizers, one paddle, and the flexible connector) are fabricated by the same origami-inspired printable method as the oscillator and thus integrated into a monolithic design. In other words, the mechanical subsystem of the printable swimmer can be directly made out of a single sheet material (Supplementary Fig. S8A). This monolithic design and fabrication strategy improves the robot's simplicity and hence, reduces the manufacturing complexity and cost.

Once the power is supplied, the paddle generates thrust on the swimmer in two steps (Fig. \ref{fig:7}C). In the first step, from state (i) to (ii), the bistable beam snaps back (leftward in the figure) driving the paddle counter-clockwise (CCW) to propel water, which results in a leftward gliding of the swimmer by a distance, $d_{ccw}$. In the second step, from state (ii) to (iii), the bistable beam snaps through (rightward in the figure) and thus the paddle rotates clockwise (CW), thrusting the swimmer leftward by another distance, $d_{cw}$. Therefore, the robot can swim leftward by $d(=d_{ccw} + d_{cw})$ in one cycle. Over time, the swimmer can continuously glide on water surface only with a constant current supply. 

We experimentally demonstrated the capability of the swimming robot, with an on-board, integrated oscillator. After a 3-second's rest, the swimmer was powered by a constant current supply and traveled 120.8 mm in 33.0 s, achieving an average speed of about 3.66 mm/s (equivalent to 1.6 body lengths/min, Fig. \ref{fig:7}D and E, Supplementary Movie S5). Meanwhile, the paddle featured an oscillatory rotational motion (in purple, Fig. \ref{fig:7}E) driven by the  reciprocating motion of the bistable beam. The swimmer glided continuously with the intermittent propulsion from the paddle thanks to the small friction on water. 
Since the oscillators are low-cost, nonmagnetic and electronic-free, the resulting printable swimmer is more applicable to tasks, such as exploration, rescue, and navigation, in extreme areas (e.g. with high magnetic field, or high radiation). 

\subsection*{Animation of an LED display}
\begin{figure}[t!]
  \centering
  \includegraphics[trim=2.5in 12.1cm 2.5in 2.5cm, clip=true,width=0.46\textwidth]{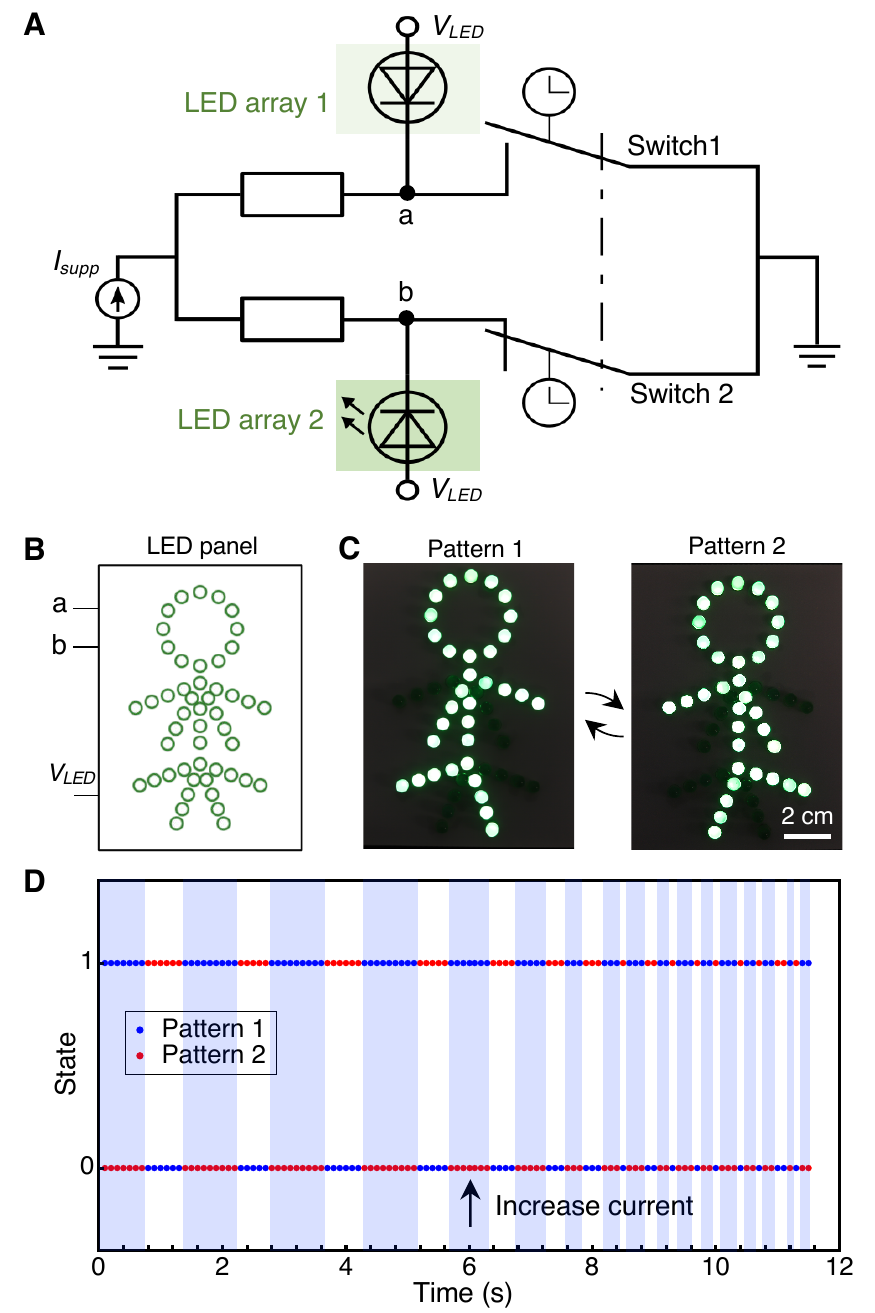}
  \caption{Animation of an LED display. Two green LED arrays are controlled by connecting their respective cathodes to points a and b of the oscillator; their power, $V_{LED}$, is independently supplied. Thus, the states of each switch can control the ``on/off” status of the corresponding LED array \textit{(A)}. The two LED arrays are integrated into a panel with a stick-figure pattern with three terminals, namely a, b, and power supply pin \textit{(B)} (see Supplementary Fig. S9 C and D for detailed circuit design). This LED panel has two different lighting pattern. When switch 1 is closed, the panel shows pattern 1; otherwise, it displays pattern 2 \textit{(C)}. \textit{(D)} Once the oscillator powered with a constant current, the LED panel (supplied with a 5.1V power source) starts to flash, alternating between pattern 1 and pattern 2. ``1" means the corresponding pattern is ``on" while ``0" represents ``off". This alternating flashing combined with the unique pattern animates the walking of the stick figure. When the amplitude of the current was increased at around 6th second, the flashing frequency grew as well, causing the stick figure to ``run'' (See Supplementary Movie S6 for the complete display). }
  \label{fig:5}
\end{figure}

The oscillator can also generate sequential electrical control: we controlled the on/off states of two LED circuits resulting in the controlled animation of the light arrays. On the basis of the oscillator, two LED arrays are incorporated into the system by connecting their cathodes at point a and b, respectively (Fig. \ref{fig:5}A). The LED light arrays can be supplied with independent power (i.e. $V_{LED}$). Therefore, the states of two LED arrays are only controlled by the open/closed status of the corresponding self-opening switches. 

Here, we integrated the two LED arrays into an LED panel composed of 49 green LED (Fig. \ref{fig:5}B, see Supplementary Note 3 for detailed fabrication specifications). This LED panel has two different light patterns (Fig. \ref{fig:5}C). When switch 1 is closed, the LED panel shows pattern 1; respectively switch 2 and pattern 2. $V_{LED}$ is about 5.1 V and is independent of the power supply of the oscillator. Once the oscillator is powered, the LED panel switched between pattern 1 and 2 leading to the designed animation of a stick figure walking. When the supply current to the oscillator was increased (at around $t$ = 6 sec), its oscillation frequency increased (Fig. \ref{fig:5}D), animating the figure to walk faster and then run (see Supplementary Movie S6 for the complete animation sequence). Due to the fabrication error, the asymmetry of the bistable beam of the oscillator led to the asymmetric oscillation, which resulted in the irregular lighting pattern in Fig. \ref{fig:5}D.
Thanks to the electronic-free nature of the oscillators, this LED flasher might work in extreme environments, such as high radiation fields, where typical electronic components could not function.

\section*{Conclusion}
Previous methods to achieve actuation and control of origami-inspired printable devices have relied on bulky and rigid components, often requiring external control system devices and limiting their potential and use in applications where completely printable robots are desirable or necessary. This paper presents a printable self-sustained compliant oscillator that is powerful, robust, configurable, and long-lasting, which can generate periodic actuation to enable driving macroscale robotic devices with prescribed autonomous behaviors. This oscillator is composed of two self-opening switches. The oscillation stems from two instabilities: the buckling instability of the bistable beam that control the ``closed" and ``opened" states of the biased contacts of each self-opening switch, and the system-level instability caused by the mechanical coupling of the poles of the two self-opening switches. This printable oscillator outputs periodic displacement and enables driving origami-inspired printable robots and mechanisms from only a single constant electrical power supply. Thus, this oscillator empowers oscillatory motions in completely printable devices, illustrating the feasibility of embedding simple control into the electromechanical constructions, eliminating the need for auxiliary electronic devices through mechanical design. In addition, a printable oscillator generating asymmetric oscillations can be obtained in a similar manner by combining switches with distinct timed delays, which enables yet additional applications. This oscillation mechanism can be characterized as a generic physical strategy to convert a constant energy input into an oscillatory output through coupled feedback of time-delayed mechanical instabilities.

The mechanism of the printable oscillator is verified through the experimental observation of the periodic displacement of the bistable beam. To enable systematic design, we built an analytical model for this printable oscillator, validating it against its constituent self-opening switch. This model allows us to predict its oscillation period against the supply current, geometric parameters, and material properties. 

We were able to achieve oscillation periods down to approximately 1.21 s by submerging the oscillator under water; the oscillation can be faster through simultaneously increasing the pulling speed and cooling speed. The snap-through motion of the beam itself provides much faster structural deformations at high force, which could be independently applicable to additional non-periodic tasks \cite{overvelde_amplifying_2015}.

We demonstrated the capability of the printable self-sustained oscillator by (i) autonomously propelling a fully printable gliding swimmer, (ii) animating an LED display, and (iii) stirring and mixing of fluid. When supplied with a constant current power, the oscillator could give rise to the aforementioned applications and others requiring periodic actuation of origami-inspired printable designs. Due to the high currents (but low voltages) necessary for the thermal actuators, a low series resistance is necessary for its power supply; with an appropriate battery, these designs can be untethered to enable infrastructureless operation. These can be put into complex environments that would otherwise preclude typical electronic components, including large ambient magnetic or radiation fields.

The oscillator is built of non-rigid raw materials through origami-inspired printable manufacturing techniques; the oscillator is preprogrammed and patterned in two-dimensional (2D) sheet material and then folded and assembled into its final three-dimensional (3D) geometry. This approach allows integrated and monolithic design and rapid fabrication, leads to accessible, low cost, and potentially disposable designs \cite{rus_design_2018}, and improves the safety for interaction with humans, making them an attractive alternative to non-printable systems that require electronic controls components.

\section*{Materials and Methods}
\subsection*{Characterization of printable devices}
The power of the devices were supplied by a laboratory DC power supply (TP-3003D-3, Kaito Electronics, Inc.). The input current and output displacement were characterized and recorded from the reading panel of the power supply and a digital camera (30 fps or 240 fps), respectively. An active cooling system was added to help speed up the cooling process of the actuator. The default parameters of the active cooling system were set to about 22$^\circ$C (room temperature), $\approx$ 110 cfm flow rate, approximately 150 mm distance from the tested devices. Other sets of parameters are available for various applications.

%%%%%%%%%%%%%%%%%%%%%%%%%%%%%%%%%%%%%%%%%%%%%%%%%%%%%%%%%%%%%%%%%%%%%%%%%%

\section*{Acknowledgements} 
We thank Mr. X. Guan for assisting in curve fitting of the analytical model of the oscillator. We also thank Mr. Y. Zhao for the help with the testing of the mechanical properties of the actuator and bistable beam. We acknowledge National Science Foundation (NSF) award no. 1644579 and no.1752575 for supporting this research.

\section*{Author contributions:}
W.Y. conceived the work, fabricated the devices, conducted the experiments, collected and analyzed data, developed the analytic model, and wrote the manuscript. A.M. revised the manuscript and oversaw the project.

\section*{Author Disclosure Statement}
W.Y. and A.M, are the inventors on a provisional patent application on the oscillator technology. The other authors declare that they have no competing interests.

%%%%%%%%%%%%%%%%%%%%%%%%%%%%%%%%%%%%%%%%%%%%%%%%%%%%%%%%%%%%%%%%%%%%%%%%%%

% Bibliography
% \bibliography{pnas-sample}

\bibliographystyle{plain}

\end{multicols}
\end{document}